\documentclass[11pt]{article}

\usepackage{float}
\usepackage{times}
\usepackage{latexsym}
\usepackage[T1]{fontenc}
\usepackage[utf8]{inputenc}
\usepackage{microtype}
\usepackage{inconsolata}
\usepackage{graphicx}
\usepackage{amsmath}
\usepackage{amssymb}
\usepackage{booktabs}
\usepackage{multirow}
\usepackage[final]{acl}

\title{Sckwoky at SHROOM-Visions: \\ Character-Level Hallucination-Span Detection \\ in Image-Conditioned VLMs via Dual-Prompted LoRA Decoders}

\author{
  Mikhail Kulakov, \ Ruslan Solikhov, \ Ivan Bondarenko \\
  Novosibirsk State University \\
  \texttt{m.kulakov@g.nsu.ru}, \texttt{r.solikhov@g.nsu.ru}, \texttt{i.bondarenko@g.nsu.ru}
}

\begin{document}
\maketitle
\begin{abstract}
We present our system for the SHROOM-visions 2026 shared task on
character-level hallucination span detection in image-conditioned
vision-language model (VLM) responses, evaluated across English,
French, Italian, and Chinese. The detector is built on a frozen
Gemma-4-12B-it backbone with trainable LoRA adapters
($\sim$10.7M parameters). Because the backbone is a causal decoder,
hidden states read from a single copy of an answer cannot attend to
subsequent tokens; we therefore repeat the candidate answer within one
prompt and extract features from its second occurrence, giving every
position approximate bidirectional context in a single
non-autoregressive pass. A character-level locator combines
intermediate hidden states to predict hallucination probabilities, BIO
boundaries, and a response-level abstention score, while separate
language-specific classifiers assign one of five hallucination types
to each predicted span. On the closed test set, our system reaches a
macro-averaged character IoU of 0.4404, more than doubling the official
baseline (0.1854), and ranks 5th--6th on IoU
across the four languages. On the English tuning split, controlled
ablations attribute a gain to the repeated answer that is comparable to
that of LoRA adaptation (its held-out gain is smaller and concentrated on
hallucinated answers), and they show that the detector relies on the
image--answer pairing: shuffling images across
examples drives IoU down to the level of the predict-nothing floor or
below it in every language. In deployments where the same Gemma-4-12B-it model
also generates answers, its base weights can be reused, making the
approach practical for local question-answering systems that cannot
run a separate verifier.
\end{abstract}

\section{Introduction}

Hallucinations in Vision-Language Models (VLMs) remain a central
reliability challenge for multimodal systems
\citep{liu2024vlmhallucinations}. The SHROOM shared-task series
introduced model-aware and model-agnostic hallucination detection
\citep{mickus-etal-2024-semeval} and was later extended to multilingual
character-level span detection in Mu-SHROOM
\citep{vazquez-etal-2025-semeval}.

\begin{figure}[htbp]
  \centering
  \includegraphics[width=\linewidth]{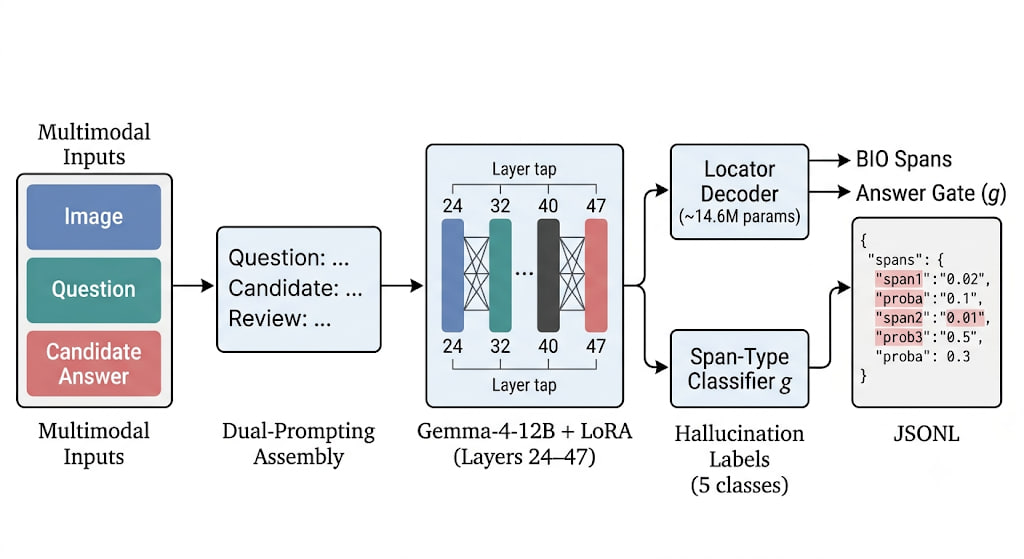}
  \caption{High-level overview of our hallucination detection pipeline, showing the information flow from multimodal input to final labeled spans.}
  \label{fig:pipeline}
\end{figure}

This paper describes our system for the SHROOM-visions 2026 shared task
\citep{vazquez2026overview}, built around the multilingual SHEEP
benchmark \citep{mickus2026sheep}. The task requires identifying
hallucinated character spans in VLM-generated answers.
Figure~\ref{fig:pipeline} gives an overview of our pipeline.

\begin{figure}[htbp]
  \centering
  \includegraphics[width=\linewidth]{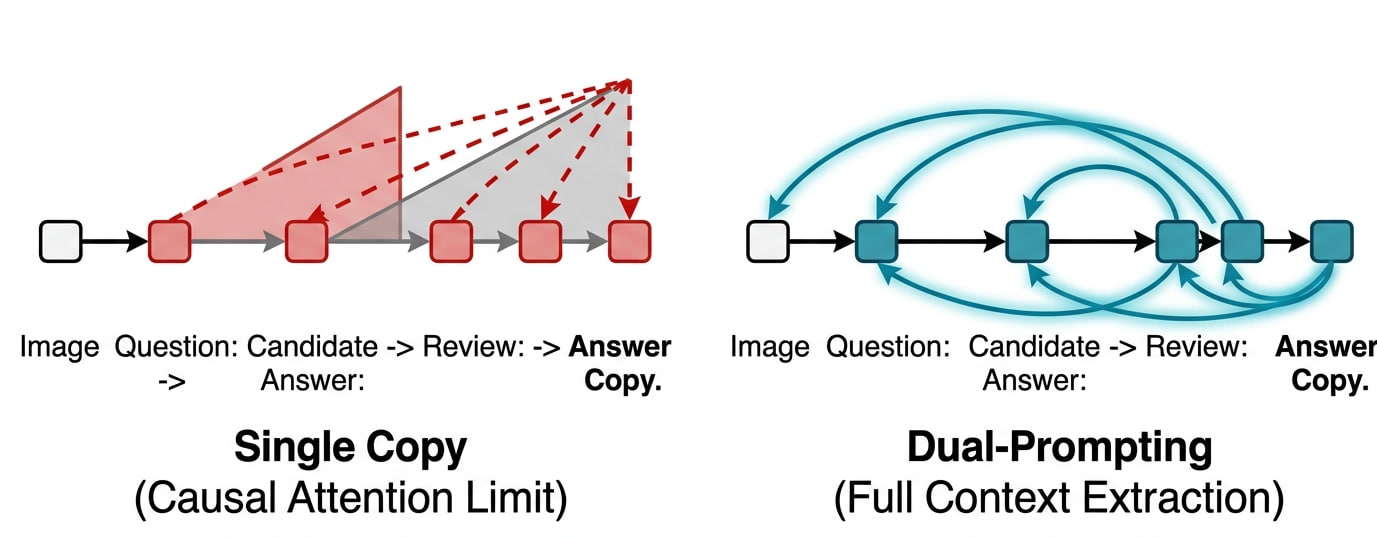}
  \caption{Conceptual visualization of our dual-prompting strategy. By duplicating the answer, tokens in the ``Review Copy'' can attend to the complete preceding copy of the candidate answer under the standard causal mask.}
  \label{fig:prompting}
\end{figure}

Our response-repetition scheme adapts echo-style contextualization, previously explored for sentence embeddings and sequence labeling, to multimodal character-level hallucination localization \citep{springer2025repetition,kukic2026sequence}. Because our backbone, Gemma-4-12B-it, is a causal decoder, hidden states extracted from a single copy of an answer cannot attend to subsequent tokens. To bypass this, our prompt includes the candidate answer twice. The first copy serves as context, while features are extracted from the tokens of the second (``review'') copy. This allows every position to attend to the image, the question, and the entire first copy (Figure~\ref{fig:prompting}), approximating bidirectional context without modifying the causal attention mask. After response generation, localization requires one non-autoregressive LoRA-enabled backbone pass over the repeated-response prompt. Span typing
is performed in a second pass through the same base checkpoint with the
LoRA adapters disabled. The implementation, evaluation scripts, and the
ablation dumps behind Section~\ref{sec:ablations} are publicly
available.\footnote{\url{https://github.com/sckwokyboom/Vision-Hallucination-Detector}}

\section{Task Setup}
\label{sec:task}

Each datapoint in SHROOM-visions 2026 consists of an image, a textual
prompt (e.g.\ a visual question), a model-generated answer, and a
character-level annotation: for every character of the answer, the
gold standard records whether it belongs to a hallucinated span and,
if so, which of five hallucination types the span belongs to. Systems
must therefore output, per character, a hallucination probability
together with a per-span type label. Evaluation is performed
separately on four languages: English, French, Italian, and Chinese.
To illustrate, an answer such as ``\textit{There are five apples and a
banana}'' to an image showing exactly three apples may yield the
spans ``\textit{five}'' (Miscounting) and ``\textit{a banana}''
(Invention).

\section{Data and Splits}

Our system relies exclusively on the official SHROOM-visions 2026
training data \citep{mickus2026sheep}, without external or synthetic
augmentation.

To ensure rigorous evaluation, we implemented a deterministic split rule based on the MD5 hash of the image name: \texttt{int(md5("13:" + image\_name)[:8], 16) / 2\textasciicircum32 < 0.1}. This ensures all answers about a single image land on the same side of the split. For English, the development side was further divided into \textit{tune-202} (for decision making) and \textit{held-out-157} (evaluated exactly once). To control cross-language leakage during multilingual training, we applied a hard filter to drop any item sharing an image with a protocol evaluation item, resulting in 12,874 training items.

\section{System Architecture}

\subsection{Backbone and LoRA}
We use \texttt{google/gemma-4-12b-it}
\citep{gemmateam2026gemma4} as our frozen backbone in
\texttt{bfloat16} precision. Images are resized to a maximum side of
768 pixels. We apply Low-Rank Adaptation (LoRA)
\citep{hu2022lora} with rank 16, $\alpha=32$, and dropout 0.05 to the
$q$, $k$, $v$, and $o$ attention projections of layers 24 through 47. This range encompasses all four layers used for feature tapping, introducing approximately 10.7M trainable parameters.

\subsection{Locator Decoder}

\begin{figure}[htbp]
  \centering
  \includegraphics[width=\columnwidth]{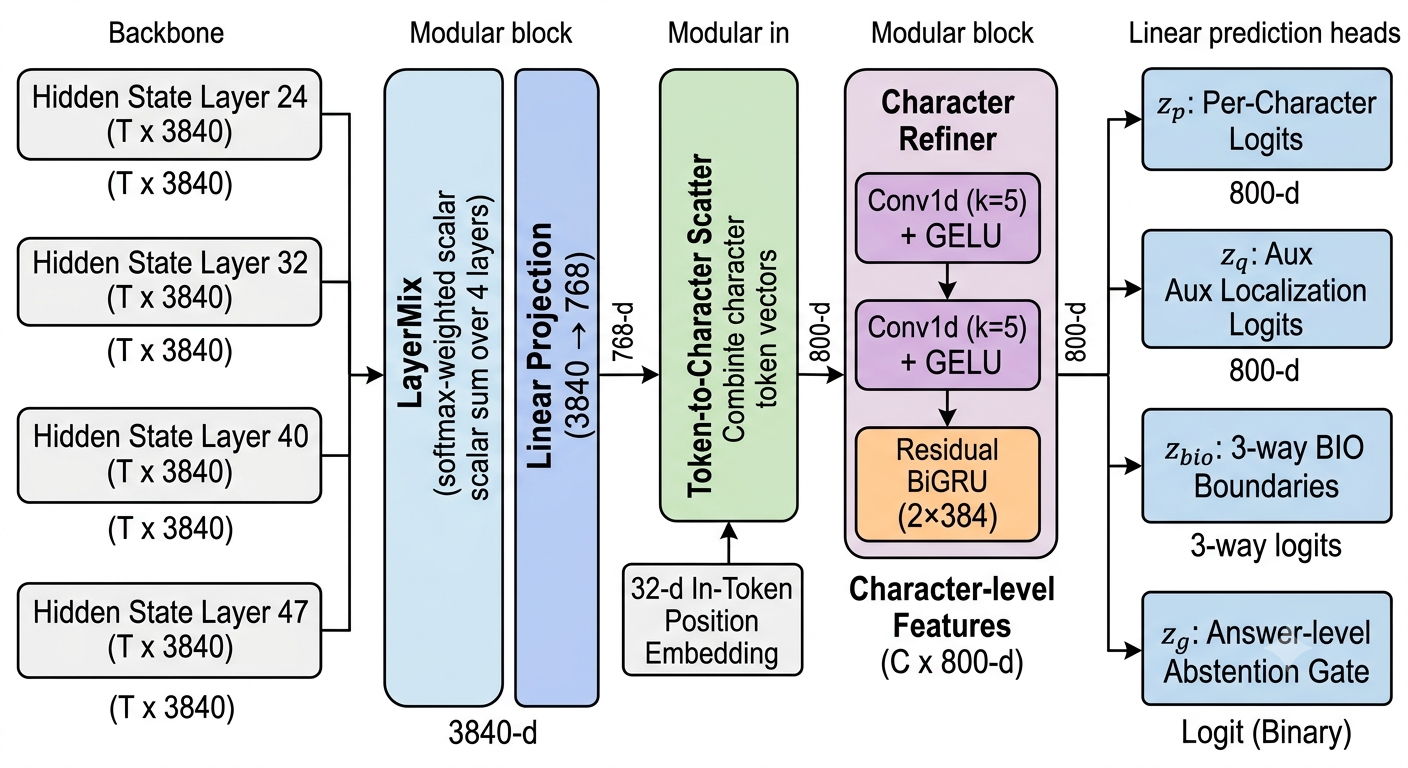}
  \caption{Detailed architecture of the Locator Decoder. The system mixes hidden states from four backbone layers, converts tokens to characters with position embeddings, and refines them via Conv1d and BiGRU stages before outputting four specialized logits.}
  \label{fig:decoder}
\end{figure}

Features $\mathbf{H} \in \mathbb{R}^{T \times 4 \times 3840}$ are extracted from layers $l \in \{24, 32, 40, 47\}$ (Figure~\ref{fig:decoder}). A \textit{LayerMix} mechanism computes a softmax-normalized weighted sum of these layers, followed by a linear projection to 768 dimensions. We use a token-to-character scatter operation: each character receives its covering token's vector concatenated with a 32-dimensional embedding of the character's position within the token. 

The character refiner consists of two Conv1d layers with GELU
activations \citep{hendrycks2016gelu}, followed by a bidirectional GRU
\citep{cho-etal-2014-learning,schuster1997bidirectional} added
residually. Four heads operate on these refined states: $z_p$
(per-character logit), $z_q$ (auxiliary localization), $z_{\mathrm{bio}}$
(O/B/I tags \citep{ramshaw-marcus-1995-text}), and $z_g$
(answer-level gate).

\subsection{Span-Type Classifier}
\label{sec:span-type}
A standalone classifier assigns one of five hallucination types to the predicted spans: \emph{Invention} (entities or content absent from the image), \emph{Mischaracterization} (incorrect description of visible content), \emph{OCR Problem} (misreading of text in the image), \emph{Miscounting} (wrong quantity of visible items), and \emph{Other}. It is trained per track on gold spans. The input features include the layer-averaged frozen hidden state of the span's tokens and three scalar length/position features. The classifier architecture is a simple multi-layer perceptron:
LayerNorm \citep{ba2016layernorm}, followed by a linear layer, GELU
activation \citep{hendrycks2016gelu}, dropout, and an output linear
layer.

\section{Training and Optimization}

The locator decoder is optimized using a composite loss function
$\mathcal{L}$ that includes a Tversky loss
\citep{salehi2017tversky}:
\begin{equation}
\begin{aligned}
\mathcal{L} &= \text{BCE}(z_p, y_{\text{soft}}) + \text{Tversky}(\sigma(z_q), m) \\
&\quad + 0.5 \cdot \text{Ranking}(z_p) \\
&\quad + 0.5 \cdot \text{CE}(z_{bio}, \text{tags}) \\
&\quad + 0.2 \cdot \text{BCE}(z_g, y_{\text{answer}}) \\
&\quad + 0.3 \cdot \text{MSE}(\sigma(z_g), \text{Top5Mean}(\sigma(z_p)))
\end{aligned}
\end{equation}
The Tversky loss ($\alpha=0.7, \beta=0.3$) heavily penalizes false negatives to optimize recall. The remaining terms are: $y_{\text{soft}}$, the soft per-character hallucination label derived from multi-annotator agreement; $m$, the binary gold hallucination mask consumed by the Tversky term; $y_{\text{answer}}$, the response-level hallucination label driving the gate; $\text{Ranking}(z_p)$, a margin loss that pushes hallucinated characters above non-hallucinated ones; and $\text{Top5Mean}(\sigma(z_p))$, the mean of the five largest per-character probabilities, used both as an abstention signal and as a self-consistency target that ties the gate to the strongest local evidence. We use AdamW \citep{loshchilov2019adamw} with constant learning rates
of $3 \times 10^{-4}$ for the decoder and $1 \times 10^{-4}$ for LoRA.

During post-processing, candidate spans are extracted via BIO argmax. Abstention is applied if $g < g_{\text{thr}}$ or $\text{Top5Mean}(p) < \tau$. Spans are trimmed, filtered for minimum length ($<3$ characters), and merged if separated by $\le 2$ characters. Finally, the external classifier relabels the spans.

\section{Results}

The task ranks systems separately per language on three character-level
metrics \citep{vazquez2026overview}: \textbf{IoU}, the
intersection-over-union between the predicted and gold sets of
hallucinated characters (span identification); \textbf{Cor}, the
correlation between the predicted per-character hallucination
probability and the empirical multi-annotator gold probability
(confidence calibration); and \textbf{Cor+Lbl}, the label-conditioned
variant of Cor in which a character contributes only when its
predicted hallucination type matches the gold type.

Table~\ref{tab:results} compares our final system with the official
SHROOM-visions baseline across the four language tracks
\citep{vazquez2026overview,shroomvisions2026web}. Our system outperformed the baseline on
all three metrics in every language, ranking 5th on IoU in
English, French, and Italian and 6th in Chinese.

Averaged across the four tracks, our system improved Cor+Lbl from
0.1854 to 0.3361, corresponding to an absolute gain of 0.1507.
(The baseline reports identical Cor+Lbl, Cor, and IoU in each track,
a property of the provided reference baseline rather than an artefact
of our evaluation.)
The average Cor score increased from 0.1854 to 0.4669
($+0.2815$), while average span IoU increased from 0.1854 to 0.4404
($+0.2550$). These results show that the improvements extend to both
confidence calibration and character-level span localization, rather
than being limited to hallucination-type classification.

Among the individual tracks, the largest Cor improvement was observed
for Italian ($+0.3291$), while French showed the largest IoU
improvement ($+0.2789$). Although the Chinese baseline was notably
stronger than the other language baselines, our system still improved
its Cor+Lbl, Cor, and IoU scores by 0.1205, 0.2328, and 0.2111,
respectively.

Classifier relabeling provided an additional improvement over our
\emph{constant-label variant}, a reference that assigns the single
most frequent hallucination type to every predicted span. Replacing
this fixed label with the learned span-type classifier
(Sec.~\ref{sec:span-type}) increased the macro-average Cor+Lbl score
from 0.2476 to 0.3361 ($+0.0885$), whereas the macro-average Cor score
changed only from 0.4631 to 0.4669 ($+0.0038$). This confirms that the
classifier primarily improved hallucination-type assignment, as
intended.

\subsection{Ablations}
\label{sec:ablations}

All ablations use the official scorer on image-disjoint development
data under the protocol of the final system: one seed (13), the best
epoch and thresholds selected on the tuning split (\textit{tune-202}
for English), and, unless stated otherwise, the submission
post-processing. English rows are also reported on
\textit{held-out-157}, which was never used for any decision.

\paragraph{Components and the repeated answer.}
Table~\ref{tab:ablations} isolates the main design choices. LoRA
adaptation was the largest single contributor (+0.039 IoU), direct
visual-token cross-attention did not help, and joint multilingual
training improved Chinese by +0.032, which mattered because the Chinese
training split contains the fewest hallucinated answers. The
\emph{single-copy} control tests the paper's central claim directly:
the same detector (LoRA, LayerMix, Conv1d\,+\,BiGRU refiner, identical
loss, schedule and seed) is trained on a prompt that contains the answer
once, so every feature is strictly causal. Under the trainer decode of
Table~\ref{tab:ablations} (before post-processing) it loses 0.029 IoU on
tune-202 (0.415 vs.\ 0.386) and 0.009 on held-out-157 (0.385 vs.\
0.376), and 0.032 / 0.010 Cor; with the submission post-processing the
IoU gaps are 0.035 and 0.006. The loss is concentrated where the
mechanism should matter: on hallucinated answers, span IoU drops from
0.326 to 0.283 (tune-202) and from 0.294 to 0.235 (held-out-157). The
character-level BiGRU therefore does not substitute for answer-level
context inside the backbone. With one seed, the held-out difference in
\emph{overall} IoU is within the run-to-run variation we measured for
the frozen baseline; the localisation gain on hallucinated answers is
consistent across both splits.

\paragraph{Does the detector use the image?}
Table~\ref{tab:image} evaluates the submitted checkpoint of every track
with the correct image, with images shuffled across examples (a seeded
derangement over the split's unique images, so no answer keeps its own
image), and with no image, at the submitted thresholds. A mismatched
image is not ignored but actively misleading: the detector flags almost
every answer (only 2--14\% of clean answers stay empty; 40\% for
Italian), and IoU falls by 0.12--0.27, to within 0.01 of the
predict-nothing floor or below it in every track. Removing the image costs 0.05--0.13 IoU and
0.05--0.12 Cor: the model then abstains more and localises worse on
hallucinated answers. Re-tuning the thresholds for the degraded
conditions recovers only part of the loss (English tune-202: 0.325
without the image vs.\ 0.422 with it). All paired image-cluster
bootstrap CIs of these differences exclude zero
(Appendix~\ref{sec:postproc}). The predictions are thus driven by the
image--answer pairing; the no-image scores above the floor show that
answer-internal cues carry some signal, but not the bulk of it.

\paragraph{Post-processing and threshold sensitivity.}
The minimum span length (3) and merge gap (2) were swept on the English
tuning split and shared across tracks. Re-running the sweep on the
ablation dumps (Table~\ref{tab:postproc}) shows that on English, French
and Italian the choice barely matters: IoU varies by at most 0.008 over
$\text{min\_len}\in\{1,2,3\}$, and by at most 0.002 over
$\text{merge\_gap}\in\{0,\dots,3\}$. Chinese is the exception the
reviewers anticipated: one CJK character can carry a complete error, and
the 3-character rule costs 0.015 IoU (0.500 vs.\ 0.515 at
$\text{min\_len}=1$) and 0.020 Cor on the Chinese development split.
Table~\ref{tab:results} keeps the submitted setting because the
evaluation phase is closed; a script-dependent minimum length is the
obvious fix. Finally, to quantify what the per-track thresholds tuned on
the French, Italian and Chinese development splits bought, we scored
those tracks with the English operating point fixed before looking at
them ($\tau=0$, $g_{\text{thr}}=0.5$, same post-processing): IoU changes
by $-0.032$ for French (95\% CI $[-0.058,-0.008]$), $-0.002$ for Italian
and $-0.014$ for Chinese (both CIs include zero), so all three tracks
remain far above their floors without any track-specific tuning.

\begin{table}[t]
\centering
\small
\setlength{\tabcolsep}{3.5pt}
\begin{tabular}{llccc}
\toprule
\textbf{Track} & \textbf{System} &
\textbf{Cor+Lbl} & \textbf{Cor} & \textbf{IoU} \\
\midrule

\multirow{2}{*}{EN}
 & Baseline & 0.1549 & 0.1549 & 0.1549 \\
 & \textbf{Our system} & \textbf{0.3129} & \textbf{0.4491} & \textbf{0.4189} \\
\addlinespace

\multirow{2}{*}{FR}
 & Baseline & 0.1679 & 0.1679 & 0.1679 \\
 & \textbf{Our system} & \textbf{0.3181} & \textbf{0.4378} & \textbf{0.4468} \\
\addlinespace

\multirow{2}{*}{IT}
 & Baseline & 0.1635 & 0.1635 & 0.1635 \\
 & \textbf{Our system} & \textbf{0.3373} & \textbf{0.4926} & \textbf{0.4293} \\
\addlinespace

\multirow{2}{*}{ZH}
 & Baseline & 0.2554 & 0.2554 & 0.2554 \\
 & \textbf{Our system} & \textbf{0.3759} & \textbf{0.4882} & \textbf{0.4665} \\
\midrule

\multirow{2}{*}{Avg.}
 & Baseline & 0.1854 & 0.1854 & 0.1854 \\
 & \textbf{Our system} & \textbf{0.3361} & \textbf{0.4669} & \textbf{0.4404} \\
\bottomrule
\end{tabular}
\caption{Official test-set comparison between our final system and the
SHROOM-visions baseline. Avg. denotes the unweighted macro-average
over the four language tracks.}
\label{tab:results}
\end{table}

\begin{table}[t]
\centering
\small
\setlength{\tabcolsep}{4pt}
\begin{tabular}{lc}
\toprule
\textbf{Configuration change (dev set)} & \textbf{$\Delta$ IoU} \\
\midrule
Add LoRA adapters (vs.\ frozen backbone) & $+0.039$ \\
Add visual-token cross-attention & $\approx 0$ \\
Joint multilingual training (ZH dev)    & $+0.032$ \\
\midrule
Single answer copy (vs.\ echo), tune-202 & $-0.029$ \\
\quad held-out-157 & $-0.009$ \\
\quad hallucinated answers only, tune-202 & $-0.043$ \\
\quad hallucinated answers only, held-out-157 & $-0.059$ \\
\bottomrule
\end{tabular}
\caption{Development-set ablations (English unless stated; trainer decode
at the tuning-split thresholds, before post-processing). Changes in IoU
are relative to the ablated configuration of each row; the multilingual
row is measured on Chinese. The single-copy rows use the same detector,
loss, schedule and seed with the answer present once in the prompt.}
\label{tab:ablations}
\end{table}

\begin{table}[t]
\centering
\footnotesize
\setlength{\tabcolsep}{2pt}
\begin{tabular}{lcccc}
\toprule
\textbf{Split} & \textbf{Floor} & \textbf{Correct} & \textbf{Shuffled} & \textbf{No image} \\
\midrule
EN tune     & 0.213 & 0.422/0.453 & 0.183/0.130 & 0.297/0.332 \\
EN held-out & 0.210 & 0.381/0.423 & 0.211/0.169 & 0.271/0.311 \\
FR dev      & 0.297 & 0.450/0.437 & 0.224/0.143 & 0.347/0.339 \\
IT dev      & 0.289 & 0.416/0.450 & 0.299/0.318 & 0.366/0.403 \\
ZH dev      & 0.355 & 0.500/0.523 & 0.225/0.223 & 0.398/0.414 \\
\bottomrule
\end{tabular}
\caption{Image dependence of the submitted checkpoints (IoU / Cor,
official scorer, full pipeline at the submitted thresholds). EN rows are
tune-202 and held-out-157; Floor is the predict-nothing IoU of the split.
\emph{Shuffled}: every answer is paired with another image;
\emph{No image}: the prompt contains no image.}
\label{tab:image}
\end{table}

\section{Conclusion}

We presented an effective, non-generative character-level hallucination detection pipeline. By utilizing a dual-prompting strategy to secure bidirectional context and decoupling boundary detection from span classification, our system achieved robust performance across all four languages in the SHROOM-visions 2026 shared task. On the English tuning split, the repeated answer contributes a gain comparable to that of LoRA adaptation, with a smaller held-out gain that is concentrated on hallucinated answers; the detector's predictions depend on the image--answer pairing rather than on textual regularities alone.

\section*{Limitations}

Our approach and experimental setup have several limitations:
\begin{itemize}
    \item \textbf{Single Seed:} Experiments, including the ablations of Section~\ref{sec:ablations}, use a single random seed (13). Run-to-run variation was estimated only for the frozen baseline ($\pm0.0060$ IoU), so differences of that size---such as the held-out-157 gap between the echo and single-copy prompts---should not be over-interpreted.
    \item \textbf{Data Leakage in Decision Making:} The \textit{tune-202} split was reused for multiple threshold sweeps and post-processing decisions. While mitigated in English via the \textit{held-out-157} split, the French, Italian, and Chinese tracks lacked untouched held-out data. Section~\ref{sec:ablations} bounds the effect: with the English operating point fixed in advance, those tracks lose between 0.002 and 0.032 IoU.
    \item \textbf{Classifier Distribution Shift:} The span-type classifier was trained on gold spans but deployed on predicted (imperfect) spans. Validation splits for this classifier were span-random rather than image-disjoint.
    \item \textbf{Language-Specific Tuning:} The minimum-span-length rule (3 characters) was tuned on English and applied uniformly. For Chinese, where a single character can encompass a severe factual hallucination, this costs 0.015 development IoU (Table~\ref{tab:postproc}); the submitted Chinese results therefore understate what a script-aware rule would achieve.
    \item \textbf{Image Dependence:} The image-shuffle and no-image controls are evaluation-time interventions on models trained with the correct image. A detector trained without images could exploit textual regularities more effectively than our no-image numbers suggest; we did not train such a model.
\end{itemize}

\bibliography{custom}

\appendix

\section{Reproducibility Details}
\label{sec:repro}

Beyond the hyperparameters given in Sections~3 and 4 (LoRA rank 16,
$\alpha=32$, dropout 0.05 on the $q,k,v,o$ projections of layers
24--47; AdamW with constant learning rates of $3{\times}10^{-4}$ for
the decoder and $1{\times}10^{-4}$ for LoRA; \texttt{bfloat16}
precision; images resized so that the longest side does not exceed
768 pixels; seed 13), the per-run batch size, epoch count, and
hardware configuration are recorded in the configuration files of our
public repository (Sec.~1). The locator decoder and the LoRA adapters
are optimized jointly, while the span-type classifier is trained
independently per track on gold spans.

\section{Ablation Details}
\label{sec:postproc}

\paragraph{Protocol.} The single-copy control was trained with the
final English recipe (cold start, seed 13, 5 epochs, LoRA on layers
24--47, decoder-only warm-up epoch) and the prompt
``\texttt{Question: \{q\}\textbackslash nCandidate answer:\textbackslash n\{a\}}'', so
that the answer's tokenisation and token-to-character alignment are
identical to the review copy of the echo prompt; its best epoch (4) and
thresholds ($\tau=0.4$, $g_{\text{thr}}=0.4$) were selected on tune-202
exactly as for the echo model ($\tau=0$, $g_{\text{thr}}=0.5$). The
image controls re-run the submitted checkpoints once per condition with
one model load; the shuffled condition uses a seeded derangement over the
unique images of the split. All post-processing and threshold sweeps
are computed offline from the dumped per-character probabilities, gate
scores and BIO tags. Uncertainty is a 95\% percentile CI from 2{,}000
bootstrap resamples of image clusters (all answers about an image are
resampled together) of the paired per-item difference.

\paragraph{Image controls.} Paired differences in IoU relative to the
correct image (Table~\ref{tab:image}): shuffled $-0.239$
$[-0.311,-0.177]$ (EN tune-202), $-0.171$ $[-0.242,-0.111]$ (EN
held-out-157), $-0.225$ $[-0.273,-0.176]$ (FR), $-0.116$
$[-0.163,-0.072]$ (IT), $-0.275$ $[-0.336,-0.214]$ (ZH); no image
$-0.125$ $[-0.185,-0.068]$, $-0.110$ $[-0.161,-0.064]$, $-0.103$
$[-0.143,-0.060]$, $-0.049$ $[-0.082,-0.019]$ and $-0.102$
$[-0.145,-0.059]$, respectively. With images shuffled, the share of
clean answers left empty drops from 0.75--0.89 to 0.02--0.14 (0.40 for
Italian); without an image it stays at 0.52--0.86 while the span IoU on
hallucinated answers falls (e.g.\ ZH 0.286 $\rightarrow$ 0.145).

\begin{table}[t]
\centering
\small
\setlength{\tabcolsep}{3.5pt}
\begin{tabular}{lccccc}
\toprule
\textbf{Split} & \textbf{1} & \textbf{2} & \textbf{3} & \textbf{4} & \textbf{5} \\
\midrule
EN tune-202     & 0.416 & 0.421 & \textbf{0.422} & 0.405 & 0.397 \\
EN held-out-157 & 0.386 & \textbf{0.388} & 0.381 & 0.373 & 0.364 \\
FR dev          & 0.448 & 0.449 & 0.450 & \textbf{0.451} & 0.440 \\
IT dev          & 0.418 & \textbf{0.419} & 0.416 & 0.415 & 0.404 \\
ZH dev          & \textbf{0.515} & 0.513 & 0.500 & 0.477 & 0.458 \\
\bottomrule
\end{tabular}
\caption{Sensitivity of IoU to the minimum span length (columns; the
submitted value is 3) at the submitted merge gap of 2. Varying the merge
gap over $\{0,\dots,3\}$ changes IoU by at most 0.002 in every track.}
\label{tab:postproc}
\end{table}

\end{document}